\documentclass[letterpaper]{article} 
\usepackage{aaai2026}  
\usepackage{times}  
\usepackage{helvet}  
\usepackage{courier}  
\usepackage[hyphens]{url}  
\usepackage{graphicx} 
\usepackage{natbib}  
\usepackage{caption} 
\usepackage{algorithm}
\usepackage{algorithmic}

\usepackage{newfloat}
\usepackage{listings}
\DeclareCaptionStyle{ruled}{labelfont=normalfont,labelsep=colon,strut=off} 
\floatstyle{ruled}
\newfloat{listing}{tb}{lst}{}
\floatname{listing}{Listing}
\usepackage{booktabs}
\title{LexChronos: An Agentic Framework for Structured Event Timeline Extraction in Indian Jurisprudence}
\author {
    Anka Chandrahas Tummepalli,
    Preethu Rose Anish
}
\affiliations {
    TCS Research\\
    ankachandrahas.t@tcs.com, preethu.rose@tcs.com
}

\usepackage{bibentry}

\begin{document}

\maketitle

\begin{abstract}
Understanding and predicting judicial outcomes demands nuanced analysis of legal documents. Traditional approaches treat judgments and proceedings as unstructured text, limiting the effectiveness of large language models (LLMs) in tasks such as summarization, argument generation, and judgment prediction. We propose \textit{LexChronos}, an agentic framework that iteratively extracts structured event timelines from Supreme Court of India judgments. \textit{LexChronos} employs a dual‐agent architecture: a LoRA‐instruct‐tuned extraction agent identifies candidate events, while a pre‐trained feedback agent scores and refines them through a confidence-driven loop. To address the scarcity of Indian legal event datasets, we construct a synthetic corpus of 2000 samples using reverse‐engineering techniques with DeepSeek-R1 and GPT-4, generating gold‐standard event annotations. Our pipeline achieves a BERT-based F1 score of 0.8751 against this synthetic ground truth. In downstream evaluations on legal text summarization, GPT-4 preferred structured timelines over unstructured baselines in 75\% of cases, demonstrating improved comprehension and reasoning in Indian jurisprudence. This work lays a foundation for future legal AI applications in the Indian context, such as precedent mapping, argument synthesis, and predictive judgment modelling, by harnessing structured representations of legal events.
\end{abstract}


\section{Keywords}
Agentic AI, Event Extraction, Legal Document Analysis, Indian Jurisprudence. 
\setcounter{secnumdepth}{1}
\section{Introduction}
\label{sec:intro}
Legal reasoning in high-stakes contexts demands not only interpretive precision but also a structured understanding of complex case narratives. This requirement underscores the critical role of event extraction—the process of identifying and structuring key facts, actors, and temporal relationships \cite{xian2024dlee}. Despite its importance, event extraction remains a persistent challenge for legal AI systems due to the inherent complexity and unstructured nature of judicial documents, particularly at the document level where tracking events across lengthy texts is essential.

Globally, AI has advanced beyond theoretical applications to practical tools in the legal sector, supporting tasks such as transcription, translation, case management \cite{pib_2025}, smart scheduling and predictive analytics for backlog reduction. Large language models (LLMs) are increasingly deployed for legal research, drafting and analysis \cite{deeks2025large}. However, extracting structured event timelines from court judgments remains largely unexplored—especially within the intricate landscape of the Indian judiciary, one of the largest in the world, with over 86,742 pending cases in the Supreme Court alone as of 2025 \cite{medianama_2025_ai_tools_sc}.

Indian legal texts pose unique challenges: dense statutory language, common law references, constitutional citations, and procedural dependencies. Traditional NLP techniques often fail to capture the temporal, causal, and hierarchical structures embedded in these documents. While sentence-level event extraction has seen progress, document-level extraction requiring co-reference resolution, entity tracking across paragraphs, and temporal-causal linkage remains a bottleneck \cite{huang2021document}. The absence of structured representations significantly limits downstream tasks such as precedent mapping, argument generation, and judgment prediction.

Although recent advances in LLMs for legal applications \cite{tiwari2024aalap} have improved text understanding, most approaches treat legal documents as monolithic text blocks, overlooking the nuanced event structures that underpin legal reasoning. Even as the Supreme Court of India adopts AI for administrative efficiency, comprehensive event extraction from judgment remains an underexplored frontier with transformative potential.

This paper addresses two key gaps: 

(1) Limitations of LLMs in extracting and structuring events from unstructured court documents. (2) the lack of publicly available event-level annotated datasets for Indian Supreme Court judgments, hindering reproducibility and benchmarking.   

To bridge these gaps, we introduce \textit{LexChronos}, an agentic framework for iterative extraction of structured event timelines from legal texts. Our dual-agent pipeline comprises: 

1. An extraction agent (LoRA-instruct-tuned LLMs $<4B$ parameters) propose candidate events, including timestamps, descriptions, judge(s) and precedent references.     

2. A feedback agent (LLMs $<4B$ parameters) evaluate candidates, assign confidence scores, and refine outputs through a stopping-criteria–driven loop. This meta-cognitive feedback ensures semantic coherence and factual accuracy, converging toward high-confidence timeline.

To establish a benchmark, we construct a synthetic corpus of 2000 Supreme Court judgments using DeepSeek-R1 \cite{guo2025deepseek} and GPT-4 \cite{achiam2023gpt}, complete with gold-standard event annotations.

Our key contributions are:     
\begin{itemize}
    \item A dual agent iterative refinement framework for extracting structured event timelines from unstructured legal texts.
    \item A synthetic, annotated dataset of 2000 Indian Supreme Court judgments tailored for even-level legal NLP tasks.
\end{itemize}

By introducing structured event timelines as an intermediate representation, \textit{LexChronos} enhances transparency and reliability in legal reasoning, paving the way for advanced tools in precedent mapping, argument generation, and judgment prediction. This work lays the foundation for intelligent and accountable AI integration in India’s judicial system.   

The rest of the paper is organized as follows: Section \ref{sec:related_work} reviews Related work, Section \ref{sec:dataset_cre} explains Dataset creation, Section \ref{sec:methodology} presents Methodology, Section \ref{sec:exp_eva} presents Experimentation and Evaluation results, Section \ref{sec:dca} provides Data and Code Availability, Section \ref{sec:tl} presents Threats to Validity and Section \ref{sec:conc} concludes with Conclusion and Future work.

\section{Related Work}
\label{sec:related_work}
Event extraction plays a pivotal role in transforming unstructured text into structured representations, enabling downstream tasks such as summarization, timeline generation, and decision support. Across domains, from biomedical literature to financial disclosures, news media, and legal documents, event extraction has evolved to address domain-specific challenges such as nested structures, long-distance dependencies, and semantic ambiguity. This section reviews key developments in three segments: event extraction across domains, specialized approaches in the legal domain, and emerging efforts tailored to the Indian legal system. 
\setcounter{secnumdepth}{2}
\subsection{Event Extraction Across Domains}
\label{sec:2.1}
In the biomedical domain, to handle nested event structures and long-distance dependencies, the BEESL model \cite{alan2020biomedical} reframed biomedical event extraction as a unified sequence labeling task using BERT-based encoding, multi-label decoding and multi-task learning enabling end-to-end handling of structural complexity.  Additionally, CNNs integrated with dependency path embeddings and diverse feature types have proven effective in pipeline-based event and relation extraction \cite{bjorne2018biomedical}.  

In the financial domain, documents such as corporate announcements present challenges due to their length and semantic density. The DocFEE  dataset \cite{chen2025dataset} supports document-level Chinese financial event extraction, while the Matrix-Chunking Method (MACK) \cite{huang2024extracting} introduces fault-tolerance mechanisms for robust extraction from raw text. Similarly, in news media, hybrid approaches combining pattern-based heuristics, machine learning models and word embeddings have been used to maintain high trigger accuracy in real-time risk assessment scenarios \cite{han2018event}.

\subsection{Event Extraction in the Legal Domain}
\label{sec:2.2}
Legal event extraction has evolved from rule-based systems to sophisticated language model-driven frameworks. Early work by \cite{sierra2018event} used part-of-speech tagging and grammatical pattern matching to extract core event attributes from Mexican Spanish legal texts. \cite{filtz2020events} benchmarked CRFs, deep learning and fine-tuned BERT variants on the ECHR dataset, revealing that CRFs excel at actor identification, while BERT variants outperform in temporal and procedural classifications. \cite{xian2024dlee} introduced DLEE, the first large-scale Chinese legal document-level event extraction dataset. They employed a semi-automated annotation pipeline combining expert-designed schemas, trigger lexicons, and a BERT-based QA model to ensure fine-grained, high-quality labeling. 

\cite{xu2024research} combined BERT, BiLSTM, and CRF architectures to extract event entities in economic legal texts, achieving robust cross-genre performance. Recent advances in LLMs have enabled more flexible and context-aware extraction. \cite{yue2024event} proposed a cooperative framework for criminal court view generation, using LLMs to construct event narratives supporting verdict explanations.

Graph-based approaches, such as those by \cite{zhao2024method}, use event entity extraction to construct judicial knowledge graphs, enhancing document interpretability and legal reasoning.  

\begin{table*}[t]
	\begin{tabular}{rlp{9cm}}\toprule
		\textit{\textbf{Judgment Component}} & \textit{\textbf{LES Coverage}} & \textit{\textbf{Justification of Coverage}} \\ \midrule
		Facts & Timestamp, Event & Factual elements are encoded as \textbf{Events} with corresponding \textbf{Timestamps} establishing the case chronology. \\
		Issues & Event & The framing of issues is a critical procedural \textbf{Event} which dictates the focus of the subsequent legal analysis. \\
		Petitioner's Arguments & Event, Precedent & Arguments constitute procedural \textbf{Events}; their basis relies heavily on \textbf{Precedent} or statutory references. \\
		Respondent's Arguments & Event, Precedent & Similar to petitioner's claims, these are timed procedural \textbf{Events} grounded in \textbf{Precedent}. \\
		Analysis of the Law & Event, Precedent & The application of law is a judicial \textbf{Event} and is intrinsically linked to the \textbf{Precedent} being discussed. \\ 
		Precedent Analysis & Event, Precedent, Judge & This analytical \textbf{Event} explicitly involves the cited \textbf{Precedent} and the authoritative \textbf{Judge} performing the analysis. \\
		Court's Reasoning & Timestamp, Event, Judge & The reasoning process forms the decisive \textbf{Event}, rendered by the \textbf{Judge}, leading up to the final \textbf{Timestamp} of the order. \\
		Conclusion & Timestamp, Event, Judge & The final order is the conclusive \textbf{Event}, delivered by the authoritative \textbf{Judge} at the final \textbf{Timestamp}. \\\bottomrule
	\end{tabular}
	\caption{Mapping core Judgment components to LES}
	\label{tab:schema_map}
\end{table*}

\subsection{Indian Legal Event Extraction} 
\label{sec:2.3}
In Indian Legal context, \cite{naik2023legal} developed a spaCy-based NER model to extract key entities such as court names, petitioners, and legal acts from Indian legal documents. \cite{kalamkar2022named} introduced a large, annotated corpus of Indian court judgments containing 46,545 legal named entities across 14 fine-grained types. They developed and released a baseline NER model trained on this dataset to support downstream legal NLP tasks like relation extraction and knowledge graph construction. \cite{tiwari2024aalap} introduced AALAP, a Mistral 7B model fine-tuned on a bespoke legal corpus for tasks such as issues generation, argument generation and event timeline construction. While outperforming GPT-3.5 on several tasks, AALAP underperformed in summarization  and constitutional Q\&A. \cite{hussain2024large} conducted a comparative evaluation of large language models, including Mistral and Gemma, for extracting judicial entities from Indian case law. They fine-tuned and tested these models to assess their effectiveness in automating structured information extraction tasks relevant to legal NLP. Systems like Legal Sarathi \cite{shivananda2024revolutionizing} integrates LLMs with machine learning algorithms to extract critical events and participants from unstructured legal documents. Its Retrieval-Augmented Generation (RAG) framework and Streamlit interface support real-time query-based retrieval and decision making.

Despite these advances, there remains a significant gap in resources tailored to the Indian judicial context. Existing works often focus on narrow event categories and lack comprehensive coverage. Crucially, no publicly available dataset captures the full spectrum of events in Indian Supreme Court documents.  

To address this, we introduce \textit{LexChronos}, an agentic framework designed to iteratively extract structured event timelines from Indian Supreme Court judgment documents. We construct a synthetic corpus of 2,000 documents using reverse-engineering techniques with DeepSeek R1 and GPT-4, generating high-quality gold-standard event annotations. \textit{LexChronos} employs a dual-agent architecture: a LoRA instruct-tuned extraction agent identifies candidate events, while a feedback agent refines them using confidence-driven stopping criteria.

\section{Dataset Creation}
\label{sec:dataset_cre}
To address the critical lack of publicly available resources for event-level annotation in the Indian legal domain, we constructed a synthetic dataset tailored to the structure and semantics of Indian Supreme Court judgments. This dataset was generated using a reverse-engineering pipeline designed to produce high-fidelity, structured representations of legal proceedings.  

The dataset creation pipeline involves three key stages: Case category selection, Event timeline generation, and Judgment text generation.  

\subsection{Case Category Selection}
\label{sec:3.1}
Case categories in the Supreme Court of India refer to distinct classifications of cases based on factors such as the legal issues involved, the subject matter of the dispute, the composition of the bench (coram), or other relevant considerations \cite{supreme2025classification}.  The dataset creation pipeline begins by randomly selecting a case category from a curated set of 25 case categories defined by the Supreme Court’s official case classification system \cite{sci_case_category}. These categories were chosen because they represent high-volume, socially impactful, and specialized domains—ranging from foundational areas like Criminal and Civil Law to emerging fields such as Cyber Law and Intellectual Property Rights.  This diversity ensures that the synthetic dataset avoids narrow algorithmic biases and reflects the procedural and topical breadth of Indian legal cases. The complete list of case categories is provided in Appendix \ref{sec:app_a}. 

\begin{table*}[t]
	\begin{tabular}{rllll}\toprule
		\textit{\textbf{Model}} & 
		\textit{\textbf{Avg Length of Judgments}} & \textit{\textbf{Avg Events per Case}} &
		\textit{\textbf{Avg Precedents per Case}} & 
		\textit{\textbf{Unique Vocab Size}}\\ \midrule
		DeepSeek-R1 & 1010.96 & 27 & 6 & 34623 \\
		GPT-4 & 552.92 & 19 & 3 & 17080 \\\bottomrule
	\end{tabular}
	\caption{Structural and linguistic comparison of DeepSeek R1 and GPT-4 judgments}
	\label{tab:judgment_comp}
\end{table*}

\begin{table*}[t]
	\begin{tabular}{p{16cm}}\toprule
		\textit{\textbf{Core Judgment Component with Judgment text}}\\ \midrule
		\textbf{\textit{Facts}}\\“The genesis of the dispute traces back to January 2020, when Proxima Inc. acquired a substantial stake in Zenith Corp, leading to disputes concerning corporate governance and the rights of minority shareholders.”\\
		\textbf{\textit{Issues}}\\“..the principal issues discussed revolved around corporate governance, the rights of minority shareholders, and the extent of judicial oversight permissible over corporate decisions.” \\
		\textbf{\textit{Petitioner's Arguments}}\\ “In February 2020, the minority shareholders of Zenith Corp alleged oppression and mismanagement by the board members of Proxima Inc., leading to the filing of a complaint…” \\
		\textbf{\textit{Respondent's Arguments}}\\ “Proxima Inc. contended that their actions were in compliance with the Companies Act, 2013, and cited the precedent in Furlong Steel Ltd. v. Cherry Steel Corp. 5 SCC 739 to support the legality of their board decisions.” \\
		\textbf{\textit{Analysis of the Law}}\\ “Upon careful consideration of the arguments and evidence presented, we find that the need to protect minority shareholders and ensure transparency in corporate governance is paramount.” \\ 
		\textbf{\textit{Precedent Analysis}}\\ “Proxima Inc., aggrieved by the injunction, appealed to the High Court in March 2022, invoking the precedent set in Bright Solar Ltd. v. Gloomy Electric Co. 9 SCC 600...” \\
		\textbf{\textit{Court's Reasoning}} \\“It appears that the NCLT’s findings were well-founded on the basis of the evidence, particularly the forensic audit report, which revealed significant financial mismanagement.” \\
		\textbf{\textit{Conclusion}} \\“In light of the foregoing, we uphold the judgment of the NCLT, affirming the restructuring of Proxima Inc.'s board and the compensation awarded to the affected minority shareholders. The appeal is therefore dismissed...” \\\bottomrule
	\end{tabular}
	\caption{Structural Component Mapping for Synthetic Judgment Text}
	\label{tab:judgement_map}
\end{table*}

\subsection{Event Timeline Generation}
\label{sec:3.2}
Once a case category is selected, the next step involves generating a structured timeline of legal events. In this section, we explain the process of event timeline generations and its validation. 

\subsubsection{Structured Event Timeline Construction}\leavevmode

For each selected case category, we use a prompt ($Prompt_E$) to generate a structured timeline of events using DeepSeek-R1 and GPT-4 
\begin{eqnarray}
{E} = Prompt_E(case \ category)
\end{eqnarray}

Each output \textit{E} is a JSON array of event objects, where each object contains four attributes: 
\begin{itemize}
    \item \textit{Timestamp}: Time or date of the event. 
    \item \textit{Event}: Narrative description of the event. 
    \item \textit{Judge}: Name(s) of the presiding judge(s), if applicable. 
    \item \textit{Precedent}: Legal precedents cited during the event.
\end{itemize}

This four-attribute schema (henceforth referred to as LexChronos Event Schema—LES) is derived through an inductive analysis of Supreme Court of India judgment documents. By examining the structural components of judicial reasoning—such as temporal progression, judicial actors, facts and cited precedents—we abstracted these into four core attributes in LES. This enables a granular, machine-readable representation of the judgment’s legal history. 

\subsubsection{LES Validation}\leavevmode

To validate the sufficiency of the LES, we map it against the eight core components typically found in Indian Supreme Court judgments, as identified in legal information retrieval systems such as IndianKanoon \cite{indiankanoon}. Table \ref{tab:schema_map} presents this mapping, demonstrating that the four-attribute schema effectively captures all substantive legal content—facts, reasoning, and conclusions required for comprehensive legal understanding. This confirms the schema’s adequacy for event-level extraction and its non-reducibility. 

\subsection{Judgment Text Generation}
\label{sec:3.3}
With the structured event timeline in place, we proceed to generate the corresponding judgment text. This step ensures that the dataset includes both structured annotations and realistic legal narratives. 
\subsubsection{Judgment Text Construction}\leavevmode

Using the structured event timeline $E$, a second prompt ($Prompt_J$) generates a synthetic judgment document: 
\begin{eqnarray}
{J} = Prompt_J(E)
\end{eqnarray}

$Prompt_J$ is guided by patterns derived from authentic judgments sourced via platforms like IndianKanoon. This ensures that the generated judgment text ($J$) emulates the linguistic complexity, structural conventions, and formatting of real Supreme Court judgments, an essential requirement for downstream legal reasoning tasks.

By repeating the entire pipeline 2,000 times, we construct the complete synthetic dataset: 
\begin{eqnarray}
D = \{ (E_i, J_i) \}_{i=1}^{2000}
\end{eqnarray}

This dataset pairs richly annotated event timelines with corresponding judgment texts across all 25 case categories. The size of 2,000 samples was chosen to ensure sufficient coverage of linguistic and structural variability across categories while maintaining computational feasibility for generation and downstream processing. DeepSeek-R1 generated 1,000 samples, while GPT-4 produced the remaining 1,000. DeepSeek-R1 was selected for its strength in structured output and logical consistency, while GPT-4 was chosen for its fluency and rhetorical fidelity in long-form legal text generation \cite{guo2025deepseek, achiam2023gpt}. Table \ref{tab:judgment_comp} compares key structural and linguistic characteristics of judgments generated by DeepSeek R1 and GPT-4. 

\subsubsection{Validation}\leavevmode

To validate the structural fidelity of the synthetic corpus, we present a detailed component breakdown of a randomly selected judgment text $J$ from the dataset.  Table \ref{tab:judgement_map} illustrates the presence and clarity of all eight core judgment components within the synthetic document. This confirms that the generative constraints embedded in  $Prompt_J$ successfully replicate authentic judicial rhetoric and organization.  

\begin{figure}[h]
\centering
\includegraphics[width=1\columnwidth]{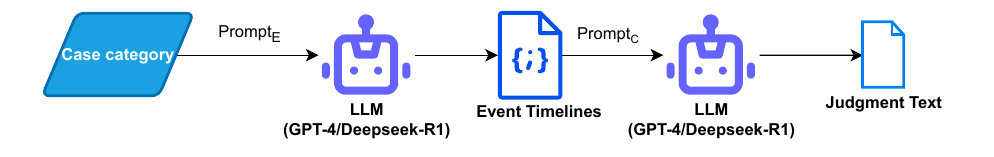} 
\caption{Reverse-Engineering Pipeline for Gold-Standard Event timeline Dataset}
\label{fig:dataset}
\end{figure}

Both $Prompt_E$ and $Prompt_J$ utilize Zero-shot prompting, enhanced with Role Prompting \cite{wang2024rolellm} and Style Prompting \cite{lu2023bounding}. These prompts are provided in the GitHub repository \cite{lexchronos2025}. This strategy ensures consistency between the structured event annotations and the generated judgment texts, functioning as a constraint satisfaction mechanism that guarantees high quality outputs. By integrating a diverse set of case categories, a rigorously defined event schema, and a dual-model generation strategy, this synthetic dataset supports robust training and evaluation of legal NLP models—particularly for event extraction tasks within the Indian legal context. Figure \ref{fig:dataset} illustrates the complete pipeline used for dataset creation. 

\section{Methodology}
\label{sec:methodology}
\textit{LexChronos}, employs a dual-agent architecture comprising an extraction agent and a feedback agent operating in an iterative refinement loop governed by predefined stopping criteria. This collaborative design ensures systematic improvement in event timeline extraction through progressive validation and feedback.

\subsection{Event Timeline Extraction}
\label{sec:4.1}
\textit{LexChronos} follows a structured, iterative process where each Judgment document $J$ undergoes progressive refinement. An instruct-tuned open-source LLM ($<= 4B$ parameters) trained via LoRA on our synthetic dataset, acts as the extraction agent. Given a Judgment document $J$, the extraction agent generates an initial set of candidate events $E_0 =(e_1 ,e_2,...,e_n)$, where each event $e_i$ includes timestamp, description, judge, and precedent information, forming a preliminary. 

A pre-trained language model ($< 4B$ parameters) serves as the feedback agent, evaluating $E_0$ against $J$ and generating feedback $F_0$ across multiple quality dimensions. The process iterates as follows: 
\begin{eqnarray}
E_0 = ExtractionAgent(J) \label{eq:initial_extraction} 
\end{eqnarray}
\begin{eqnarray}
F_0 = FeedbackAgent(J, E_0) \label{eq:initial_feedback}
\end{eqnarray}
\begin{eqnarray}
E_{i+1} = ExtractionAgent(J,E_i, F_i)
\end{eqnarray}
\begin{eqnarray}
F_{i+1} = FeedbackAgent(J, E_{i+1})
\end{eqnarray}
where $E_i$, $F_i$ denotes intermediate stages of refinement process. $E_{i+1}$ and $F_{i+1}$ will be last pair of refinement steps where $E_{i+1}$ will be the final extracted event timeline if the stopping criteria is met.   

The iterative process terminates when either of the two stopping conditions (Patience limit and tolerance threshold limit) is satisfied. 
\begin{itemize}
    \item The \textbf{patience limit} is triggered when the confidence score fails to exceed the best achieved score $S_best$ for 3 consecutive iterations: 
      \begin{eqnarray}
      \forall j \in \{i-2, i-1, i\} : S_j \le S_{best}
      \end{eqnarray}
    \item The \textbf{tolerance threshold limit} activates when confidence scores remain identical for 3 successive iterations:
  \begin{eqnarray}
  S_{i-2} = S_{i-1} = S_i
  \end{eqnarray}
\end{itemize}

Empirical evaluation showed three iterations provided the best trade-off between performance and computational cost. This process results in a refined event timeline $E_k$ accompanied by final feedback $F_k$, ensuring systematic quality improvement through collaborative agent interaction. 

\begin{figure}[h]
\centering
\includegraphics[width=1\columnwidth]{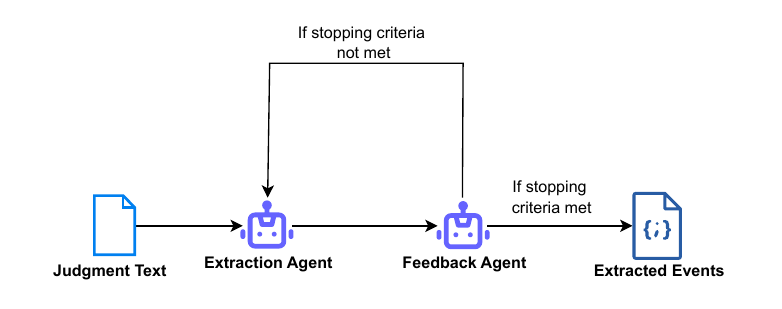} 
\caption{The LexChronos Framework for Iterative Extraction and Feedback-Driven Refinement}
\label{fig:methodology}
\end{figure}

The feedback agent scores extracted events on seven dimensions ($0.00-1.00$ scale):  \textit{Narrative Relevance} (alignment with case story), \textit{Temporal Accuracy} (correctness of dates), \textit{Chronological Flow} (logical sequencing), \textit{Event Detail} (adequacy of descriptive details), \textit{Repetition} (detection of duplicates), \textit{Character Identification} (clarity of participant roles), and \textit{Confidence Score} (overall extraction quality). A textual critique accompanies scores, highlighting improvement areas. The Confidence Score serves as the primary convergence signal. This systematic thresholding signals when the output quality has maximized its precision and recall, allowing the system to halt confidently rather than continuing costly, minimally productive generations. This structured iterative mechanism ensures that the final extracted timeline, $E_k$, possesses the highest possible verifiable reliability before being passed to downstream applications. For initial extraction and refinement, we used Zero-shot prompting with Role and Style prompting. For feedback generation, we used Zero-shot prompting with self-criticism \cite{huang2023large} and linear scale \cite{liu2023evaluate} techniques. The overall framework architecture and iterative refinement process are illustrated in Figure \ref{fig:methodology}. These prompts are provided in the GitHub repository \cite{lexchronos2025}. 

\subsection{Downstream Task Evaluation}
\label{sec:4.2}
To demonstrate the practical utility of structured event timeline representations in legal AI applications, we evaluate its impact on a critical downstream task: judgment summarization. We randomly select 200 sample judgments from the synthetic dataset introduced in Section \ref{sec:dataset_cre}, denoted as $J=(J_1,J_2,...,J_{200})$, each paired with its corresponding ground-truth event timelines $E=(E_1,E_2,...,E_{200})$. 

We conduct comparative summarization experiments using two input representations.  

\begin{itemize}
    \item Unstructured approach: Original judgment documents as provided as input requiring the LLM to extract key arguments and outcomes from verbose, noisy text.  

      \begin{eqnarray}
      S_{unstructured} = LLM(J_j) \label{eq:unstructured}
      \end{eqnarray}
    \item Structured approach: Generated event timelines serve as input, offering a concise, temporally ordered narrative. 

      \begin{eqnarray}
      S_{structured} = LLM(E_j) \label{eq:structured}
      \end{eqnarray}
\end{itemize}

Both approaches employ identical open-source language models for summary generation, ensuring fairness and isolating the contribution of structured representations of summarization quality. Summarization prompts use Zero-shot prompting combined with Role and Style prompting. We adopt zero-shot prompting for summary generation to ensure fairness and reproducibility across models, isolating the effect of input representations (structured vs unstructured) from model adaptation. While few-shot or fine-tuned models may improve performance, our objective is to benchmark the contribution of structured representations independently of adaptation effects, making zero-shot prompting the most appropriate setting for controlled evaluation. 

For evaluation, we adopt GPT-4 as the judge LLM, and employ a pairwise comparison protocol, where the evaluator receives the original judgment text alongside two candidate summaries and determines which better satisfies legal quality standards. GPT-4 was selected due to its demonstrated strong alignment with human evaluators in summarization assessment, with pairwise comparison identified as the most effective evaluation method \cite{liu2024benchmarking}. Given this high correlation, human evaluation was not included, as GPT-4 offers a reliable and cost-effective alternative for large scale experiments.  

Our evaluation protocol is guided by a detailed checklist comprising eight equally weighted criteria—\textit{Streamlined Narrative} (clear legal storyline), \textit{Conciseness} (compact yet complete), \textit{Outcome clarity} (explicit final decision), \textit{Focus on key elements} (highlights core issues), \textit{Citation economy} (minimal necessary citations to avoid legal jargon), \textit{Readability for practitioners} (plain professional language), \textit{Balanced detail} (essential context only), \textit{Chronological efficiency} (logical sequencing). These criteria collectively capture the balance between accuracy, readability and practical utility required for legal summaries. The Judge LLM selects the summary that satisfies the majority of these criteria while maintaining fidelity to the original judgment text $J$. To mitigate bias, the evaluation prompt incorporates zero-shot prompting with self-criticism, requiring GPT-4 to justify its preference with an interpretable rationale. All summarization and evaluation prompts are publicly available in our GitHub repository \cite{lexchronos2025}. 

\begin{table*}[t]
\centering
	\begin{tabular}{rlll}\toprule
		\textit{\textbf{LLM}} & 
		\textit{\textbf{Bert Precision}} & \textit{\textbf{Bert Recall}} & \textit{\textbf{Bert F1}} \\ \midrule
		Llama 3.2 3B Instruct & \textbf{0.8450} & \textbf{0.8336} & \textbf{0.8390}\\
		Gemma 3 1B IT & 0.8237 & 0.7992 & 0.8113 \\
		Gemma 2 2B IT & 0.7809 & 0.7957 & 0.7880 \\
		Llama 3.2 1B Instruct & 0.6058 & 0.6054 & 0.6052 \\
		DeepSeek R1 Distill Qwen 1.5B & 0.5548 & 0.5135 & 0.5276 \\ 
		Qwen 2.5 3B & 0.5628 & 0.4467 & 0.4752 \\
		Phi 4 Mini Reasoning & 0.4802 & 0.2442 & 0.3048 \\
		Qwen 3 4B & 0.4472 & 0.2506 & 0.3005 \\\bottomrule
	\end{tabular}
	\caption{Extraction Agent performance on $n_{test}$}
	\label{tab:finetuning_results}
\end{table*}

\begin{table}[t]
	\begin{tabular}{rp{1cm}p{1cm}p{0.9cm}}\toprule
		\textit{\textbf{Feedback Configuration}} & \textit{\textbf{Bert \newline Precision}} & \textit{\textbf{Bert Recall}} & \textit{\textbf{Bert F1}} \\ \midrule
		Self Feedback & 0.8681 & 0.8371 & 0.8523 \\
		Gemma 2 2B IT & \textbf{0.8918} & \textbf{0.8590} & \textbf{0.8751} \\
		Gemma 3 1B IT & 0.8664 & 0.8348 & 0.8502 \\
		Llama 3.2 3B Instruct & 0.8665 & 0.8351 & 0.8505\\
		Llama 3.2 1B Instruct & 0.8661 & 0.8344 & 0.8499 \\\bottomrule
	\end{tabular}
	\caption{Feedback Agent Impact on Extraction Quality}
	\label{tab:feedback_results}
\end{table}

\begin{table}[t]
	\begin{tabular}{rll}\toprule
        \textit{\textbf{LLM}} & {\boldmath$S_{\textbf{\textit{unstructured}}}$} & {\boldmath$S_{\textbf{\textit{structured}}}$} \\ \midrule
		Llama 3.1 8B Instruct  & 55 & 145\\
		Gemma 2 9B IT & \textbf{50} & \textbf{150} \\\bottomrule
	\end{tabular}
	\caption{GPT-4 Preference for Structured Input Summaries ($n_{sum}$) }
	\label{tab:summary_results}
\end{table}

\section{Experiments and Evaluation Results}
\label{sec:exp_eva}
We evaluate \textit{LexChronos} in three sequential stages: (i) instruct-tuning the extraction agent, (ii) assessing the feedback agent’s contribution to extraction quality, and (iii) measuring the impact of structured event timelines on legal judgment summarization. All experiments leverage the synthetic dataset described in Section \ref{sec:dataset_cre}. 

\subsection{Extraction Agent Performance}
\label{sec:5.1}
To identify the most effective extraction agent, we instruct-tuned eight open-source language models (each under $4$ billion parameters) on the training split of our synthetic corpus ($n_{train}=1600$), reserving $n_{test}=400$ for evaluation. All models were fine-tuned using LoRA-based instruct-tuning, which updates only a small fraction of parameters, ensuring computational efficiency. Candidate models included: Llama 3.2 3B Instruct \cite{llama3.2-3b-instruct}, Llama 3.2 1B Instruct \cite{llama3.2-1b-instruct}, Gemma 2 2B IT \cite{gemma_2024}, Gemma 3 1B IT \cite{team2025gemma}, DeepSeek R1 Distill Qwen 1.5B \cite{guo2025deepseek}, Phi-4-Mini-Reasoning \cite{xu2025phi}, Qwen 2.5 3B \cite{qwen2.5}, and Qwen 3 4B \cite{yang2025qwen3}. 

Post tuning, each model was evaluated on the held-out test set using BERT-based Precision, Recall, and F1 \cite{zhang2019bertscore} against the gold-standard event annotations {$E_i$}. The choice of BERT-based metrics is critical in the legal domain: traditional metrics such as ROUGE or BLEU rely on lexical overlap, which often fails to capture semantic nuances in legal language, where subtle word choices carry significant contextual and legal meaning \cite{kaster2021global}. BERTScore, by leveraging contextual embeddings, measures semantic coherence and equivalence between extracted event and ground truth, offering a stronger correlation with human judgment of textual fidelity. For complex, long-form legal documents, semantic equivalence is a far more reliable indicator of extraction quality than token-level matching, making BERT-based metrics are preferred choice.  

Table \ref{tab:finetuning_results} presents the results, with Llama 3.2 3B Instruct achieving the highest BERT-based F1 score (0.8390) among all candidates. This superior performance underscores that architectural designs optimized for reasoning despite smaller parameter counts significantly enhance structured information extraction. Consequently, Llama 3.2 3B Instruct was selected as the final extraction agent.

\subsection{Feedback Agent Contribution}
\label{sec:5.2}

Building on the best-performing extraction agent, we evaluate the role of the feedback agent by integrating multiple feedback models into an iterative refinement loop, with instruct-tuned Llama 3.2 3B Instruct serving as the extraction agent. We compare five configurations: (1) self-feedback \cite{madaan2023self} using instruct-tuned Llama 3.2 3B Instruct for both extraction and critique; (2)Llama 3.2 3B Instruct; (3) Llama 3.2 1B Instruct; (4) Gemma 2 2B IT; and (5) Gemma 3 1B IT. The latter four feedback agents are pre-trained LLMs without additional tuning. For each configuration, we perform up to 10 refinement iterations per judgment, applying the stopping criteria of patience limit and tolerance threshold as described in Section \ref{sec:4.1}. Table \ref{tab:feedback_results} reports BERT-based Precision, Recall, and F1 on the same test set ($n_{test}$). 

\subsection{Impact on Summarization}
\label{sec:5.3}
Building on our extraction-feedback pipeline, we evaluate the downstream benefits of structured event timelines for legal summarization. From the test set, we randomly sampled 200 judgments ($n_{sum}=200$) and generated two summaries per judgment using two open-source LLMs: Llama 3.1 8B Instruct \cite{grattafiori2024llama} and Gemma 2 9B IT \cite{gemma_2024}. For each judgment $J_j$ we produced:

\begin{itemize}
    \item Unstructured summary: as per equation \ref{eq:unstructured} from section \ref{sec:4.2} with $J_j$ as input
    \item Structured summary: as per equation \ref{eq:structured} from section \ref{sec:4.2} with $E_j$ as input, where $E_j$ denotes the ground-truth event timeline.
\end{itemize}

To assess summary quality, we employ GPT-4 as an automated evaluator. For each judgment, GPT-4 is given $J_j$,  $S_{unstructured}$, and $S_{structured}$, and asked to select the summary that best captures key information, preserves legal context, and aligns with the original document’s intent.
\begin{equation}
Best \ Summary = GPT\ 4(J_j, S_{unstructured}, S_{structured})
\label{eq:best_summary}
\end{equation}

As shown in Table \ref{tab:summary_results}, structured event timelines markedly improve summarization quality across both LLM architectures, with Gemma 2 9B IT exhibiting the highest preference (75\%). This strong preference demonstrates the transferability and utility of structured timelines across different foundational models (Meta vs. Google architectures). Our findings highlight that the extraction process acts as a knowledge pre-processor, creating a high-quality bottleneck. By supplying the summarization models with a clean, temporally ordered legal narrative, we enable them to focus on linguistic generation and high-level reasoning, rather than exhaustive document comprehension and filtering of verbose legal text.  

Collectively, the extraction-feedback evaluations validate the effectiveness of our dual-agent framework in producing high-quality event timelines, while the summarization experiments demonstrate the downstream advantages of structured representations in legal AI tasks.

\section{Data and Code Availability}
\label{sec:dca}
The code, prompts, hyper-parameters, instruct-tuned model adapters from section \ref{sec:5.1}, and the complete synthetic dataset are available in GitHub \cite{lexchronos2025}. All resources are publicly accessible.

\section{Threats to Validity}
\label{sec:tl}
While LexChronos demonstrates promising results, certain factors may influence the generalizability of our findings. (1) The synthetic dataset, although designed to emulate the structural and rhetorical characteristics of Supreme Court judgments, is not derived from actual cases. The absence of authentic judicial complexity such as rare procedural anomalies and unpredictable factual patterns may limit its applicability to real-world scenarios. (2) The dataset covers only 25 case categories out of 48 from the Supreme Court’s classification framework. While these categories represent high-volume and socially significant domains, excluding others may introduce topical bias and constrain the framework’s coverage across the full spectrum of Indian jurisprudence. (3) The current evaluation is restricted to English-language judgments. Given the linguistic diversity of India’s judicial ecosystem and the prevalence of vernacular submissions, this limitation may affect the adaptability of the framework in multilingual contexts.

\section{Conclusion and Future work}
\label{sec:conc}
This work introduces and validates \textit{LexChronos}, an agentic AI framework for extracting structured event timelines from Indian Supreme Court judgments. The proposed dual-agent architecture, combining an instruct-tuned extraction agent with a pre-trained feedback agent, offers a systematic approach to iterative output refinement. To address the lack of resources in this domain, we created a synthetic corpus of 2,000 annotated judgments, establishing a foundational benchmark for Indian legal event extraction. The best-performing configuration achieved a BERT-based F1 score of 0.8751, demonstrating high accuracy. Furthermore, downstream summarization experiments confirm the practical utility of structured timelines. Summaries generated from event-based inputs were preferred by GPT-4 in the majority of cases. These findings underscore the importance of moving beyond traditional unstructured approaches and adopting structured representations for legal AI applications. 

Looking ahead, several promising research directions emerge. A critical next step is transitioning from synthetic to real-world datasets. While the synthetic corpus enabled initial development and evaluation, a human-annotated Supreme Court dataset would validate the framework's robustness in a practical setting and serve as a valuable resource for the legal AI community. Additionally, expanding \textit{LexChronos} to cover all levels of Indian courts, including High courts and lower tribunals, is essential for achieving generalizability. Addressing linguistic diversity by supporting vernacular languages will further enhance applicability across India’s judicial ecosystem.

Beyond data, the framework can be extended to enable advanced downstream tasks. While summarization served as an initial proof of concept, future work will explore applications such as precedent mapping, enabling AI systems to trace the legal history and influence of judgments; argument generation and judgment prediction, which require modelling logical and causal relationships between events; and knowledge base construction, leveraging temporal relationships to transform raw legal text into interconnected networks of legal knowledge. These directions represent a shift from basic information extraction toward task-oriented legal reasoning, positioning \textit{LexChronos} as a foundation for next-generation legal AI systems.

\appendix
\section{List of Case Categories}
\label{sec:app_a}
Table \ref{tab:domain_areas} lists the curated set of 25 case categories considered for dataset creation, as referenced in Section \ref{sec:dataset_cre}.

\begin{table}[h]
\centering
\begin{tabular}{ll}
\toprule
\multicolumn{2}{c}{\textbf{Case Categories}} \\ \midrule
Constitutional Law & Civil Law \\
Banking and Finance Law & Education Law \\
Intellectual Property Rights (IPR) & Taxation Law \\
Labor and Employment Law & Family Law \\
Consumer Protection Law & Cyber Law \\
Real Estate and Property Law & Criminal Law \\
Service Law (Government Employees) & Contract Law \\
Environmental Law & Corporate Law \\
Administrative Law & Insurance Law \\
Health and Medical Law & Maritime Law \\
Human Rights Law & Election Law \\ 
Energy and Mining Law & Telecom Law \\
Customs and Excise Law & \\
\bottomrule
\end{tabular}
\caption{List of Case Categories}
\label{tab:domain_areas}
\end{table}

\bibliography{aaai2026}

\end{document}